\documentclass[runningheads]{llncs}

\usepackage{cite}
\usepackage[T1]{fontenc}
\usepackage{graphicx}
\usepackage{hyperref}
\usepackage{color}

\begin{document}
\title{Few-shot Class-incremental Learning for
Cross-domain Disease Classification}

\author{Hao Yang\inst{1,2,3}\and Weijian Huang\inst{1,2,3}\and  Jiarun Liu\inst{1,2,3} \and Cheng Li\inst{1} \and Shanshan Wang\inst{1,3,*}}
\authorrunning{H. Yang et al.}
\titlerunning{Few-shot Class-incremental Learning for Cross-domain Disease Classification}
\institute{Paul C. Lauterbur Research Center for Biomedical Imaging, Shenzhen Institute of Advanced Technology, Chinese Academy of Sciences, Shenzhen, Guangdong, China \and University of Chinese Academy of Sciences, Beijing, China
\and Pengcheng Laboratory, Shenzhen, Guangdong, China
\\
\email{*Corresponding:Sophiasswang@hotmail.com, ss.wang@siat.ac.cn}}

\maketitle              
\begin{abstract}
The ability to incrementally learn new classes from limited samples is crucial to the development of artificial intelligence systems for real clinical application. Although existing incremental learning techniques have attempted to address this issue, they still struggle with only few labeled data, particularly when the samples are from varied domains. In this paper, we explore the cross-domain few-shot incremental learning (CDFSCIL) problem. CDFSCIL requires models to learn new classes from very few labeled samples incrementally, and the new classes may be vastly different from the target space. To counteract this difficulty, we propose a cross-domain enhancement constraint and cross-domain data augmentation method. Experiments on MedMNIST show that the classification performance of this method is better than other similar incremental learning methods. 

\end{abstract}

\section{Introduction}
With the increase of model scale and the number of training data, deep learning has achieved better and better performance in downstream tasks. However, it is well known that a well trained deep neural network (DNN) with large scale of parameters is usually difficult to adapted to a new task by training on just a few examples. Especially when the new samples are coming from different source domains. Further, lack of ability to preserve previous knowledge limits the application of DNN \cite{goodfellow2013empirical, mccloskey1989catastrophic}. Therefore, it is important to explore the few-shot learning and memorizing capability of deep learning models.

\begin{figure}[t]
\centering
\includegraphics[width=0.8\textwidth]{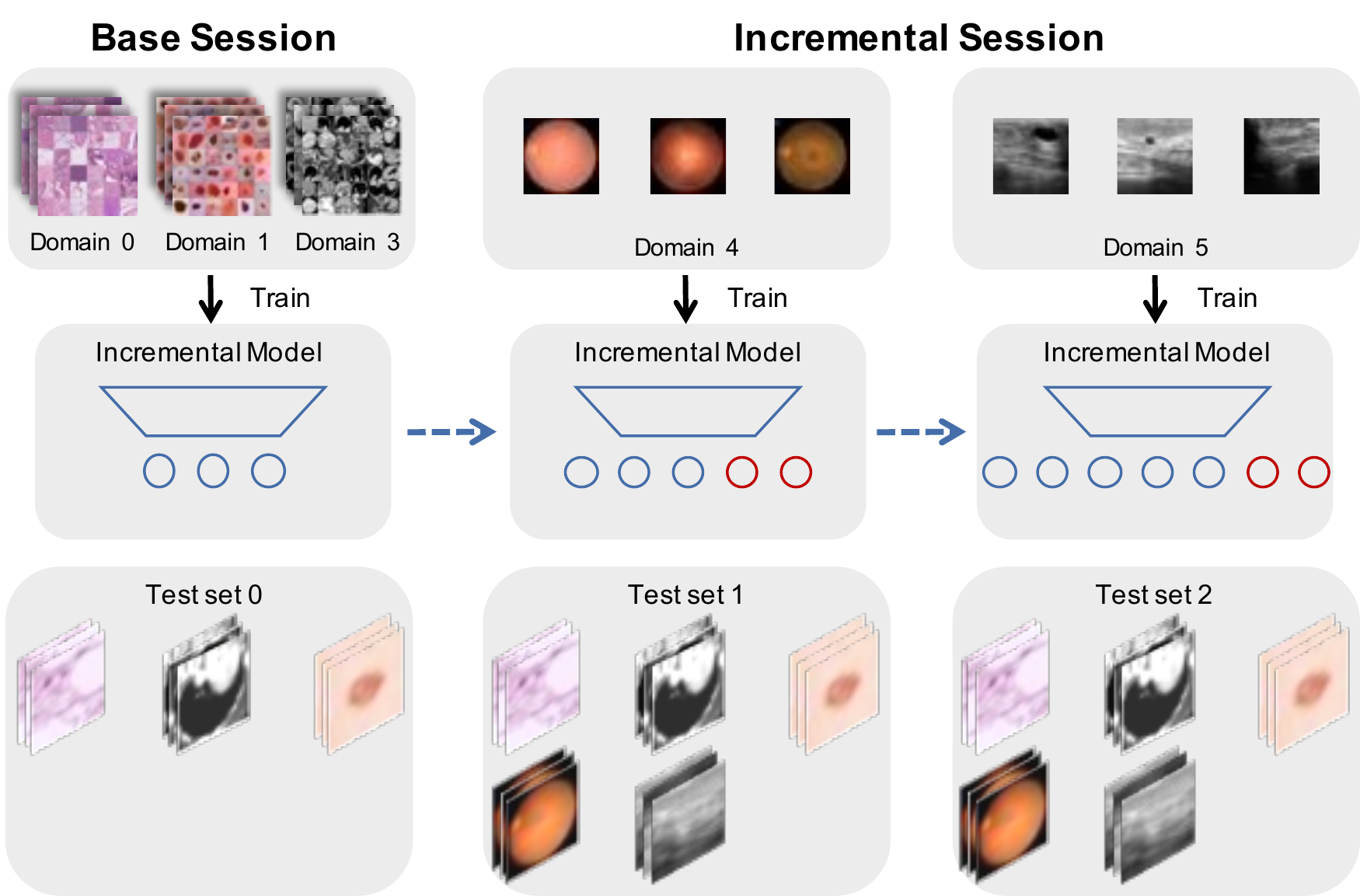}
\caption{Scenario of cross-domain few shots incremental learning. Each scenario achieved by adding non-overlapping classes sequentially. Only a few with different domain images are used for training in the incremental session, while the target space may be very different from the base session.  An ideal model should perform well in all categories after training.}
\label{fig:workflow}
\end{figure}

However, existing methods may suffer from few samples increment for which may lead to overfit of the representation and thus destroy the pre-trained structure of feature embeding\cite{finn2017model}. As results, leading to more serious forgetfulness. Reasearchers have trying to solve the few-shot incremental problem. Some methods update the backbone when receiving the new incremental tasks after training the network \cite{zhao2021mgsvf,dong2021few}, while other methods freeze\cite{zhang2021few,zhu2021self}. At the same experiment setting, comparing to the one freezing the backbone, the updated methods showing more performance decrease since the large unbalancing between the coming data and source data, especially when the data come from different domain.

To this aim, we are the first to propose a challenging and practical novel scenarios: cross-domain few-shot incremental learning (CDFSCIL). CDFSCIL requires the model to constantly adapt to new tasks by receiving data from different source domains, and retain the previously learned tasks without additional cost. Considering clinical application, CDFSCIL has the following attribution: 1) The model needs to perform well on all classes equally, even when there are a huge gap in representation space between different domain data; 2) For extreme few sample case, the model should still maintain stable performance. 

The strategy of backbone freezing decouples the learning of representation and classifier to avoid over-fitting and catastrophic forgetting in representation\cite{zhang2021few,peng2022few,limit,hersche2022constrained}. In addition, since the basic features of different objects are similar, it is meaningful to choose these types of features to identify new classes. Inspired by the decouples strategy, our method reserve space for future cross-domain tasks by constructing pseudo-domains and pseudo-labels in advance and hence lead to better class-incremental performance.

This article proposes a cross-domain few-shot class incremental learning method designed to learn more universal feature embeddings from a small amount of data. Specifically, we generate a large number of pseudo-data and train the model on it to adapt to these data. Through introducing a domain-constrained loss, we jointly optimize the pseudo-data and real data, extract invariant information and enable the model to handle incremental sessions with few samples. Our contributions are as follow: 
\begin{enumerate}
  \item We are the first to propose the challenging senario of CDFSCIL;
  \item We narrowed the cross-domain feature distribution to maximizes the distance between inter-class feature vectors via a novel constraint;
  \item We propose a new cross-domain augmentation strategy to compact intra-class clustering and wide inter-class separation.
\end{enumerate}

\begin{figure}[t]
\includegraphics[width=\textwidth]{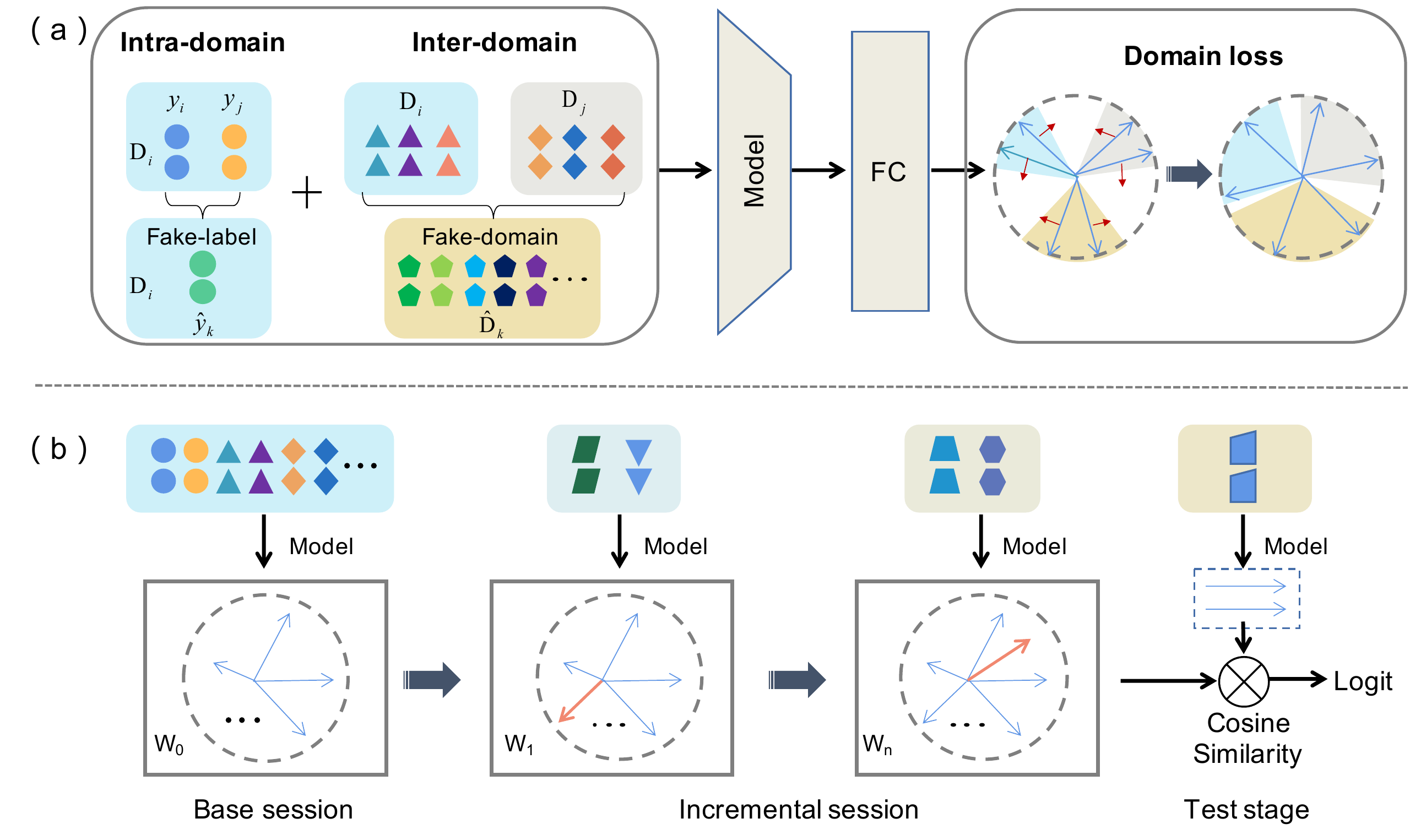}
\caption{This is the workflow of the proposed method. We generate pseudo-incremental data on the base session with multiple domains, obtaining a universal feature embedding representation by making intra-class clustering more compact and inter-class margin of the same domain wider. a) The training process on the base session first randomly samples pseudo data to simulate the future incremental phase to be learned, using domain-enhanced loss to enhance the discriminability of domain-related categories. b) The incremental learning and inference process. The newly added classifier for each session is the average embedding of each class training data in that session. The logit is obtained by the cosine similarity between the data embedding and the classifier.} \label{fig2}
\end{figure}

\section{Method}
In this section, we introduce the propose method via: 1) a cross-domain enhancement constraint and 2) intra-/inter- class separation strategy for the challenges of FSCIL. The workflow of the method are as shown in Fig. \ref{fig:workflow}.

\subsection{Problem Formulation}
The goal of FSCIL is to learn new classes from extremely limited training samples while maintaining performance over old sessions. Typically, FSCIL has several sequential learning sessions $\{S^0,S^1,...,S^B\}$. Each session contains $N^b$ labeled sample pair $S^b=\{x_i,y_i\}^{N^b}_{i=1}$, where $x_i \in R^D$ is input data and $y \in Y^b$. $Y^b$ is the label space of $S^b$ and $Y^b$ satisfying $\forall b \not= b', Y^b \cap Y^{b'} = \empty$. The training set of $b$-th session $S^i_{train}$ contains data of current session only and will be inaccessible during other sessions. Meanwhile, the testing set of $b$-th session $S^b_{test}$ includes both test sets of previous and current sessions, that is $Y^b_{test}=\{Y^0 \cap Y^1,...,Y^b\}$. Usually, the training set $S^0_{train}$ of the first session is a relatively large data set, which is also known as the basic training set. Instead, the data set is usually described as the N-way K-shot training set in all subsequent sessions. State-of-the-art continuous learning solutions impose constraints on input space and target space, based on the assumption that all tasks have the same target space \cite{chaudhry2018riemannian,lenga2020continual,memmel2021adversarial} or all tasks originate from the same (partial) data set \cite{chaudhry2018riemannian,derakhshani2021kernel}. In practice, $S$ may come from multiple different domains with large distribution discrepancies. Therefore, the goal of our CDFSCIL is continuous learning in different domains. Each domain is defined as a single data set of one class or multiple categories, and each S contains the data of multiple domains. 

\subsection{Domain enhancement constraint}
In the case of cross-domain incremental, due to the large inter-domain differences, the learning based on cross-entropy loss may lead the model to converge only in the inter-domain that are easy to classify. The attention to the intra-domain fine category is thus weakened. Therefore, we hope to use a loss function, which can not only maintain the performance of intra-domain classification but also enhance inter-class attention.

Many adaptive cross-entropy and cosine-similarity constraint have been used in classification problems \cite{liu2016large,liu2017sphereface,wang2018cosface}:  The cross-entropy constraint classify different items by maximizing the posterior probability of the groundtruth.  Given the input representation $x_{i}$ with its corresponding label $y_{i}$, the cross-entropy loss can be formulated as:
\begin{equation}
\label{eq:1}
L_{s} = \frac{1}{N}\sum_{i=1}^{N}(-logp_{i}) = \frac{1}{N}\sum_{i=1}^{N}-log\frac{e^{f_{y_{i}}}}{\sum_{j=1}^{C}e^{f_{j}}}
\end{equation}
where $p_{i}$ indicates the posterior probability, $N$,$C$ is the number of samples and classes respectively. $f_{j}$ is usually represented as the activation of the fully connected layer with the weight $W_{j}$ and the bias $b_{j}$. For simplicity, we set and fix the $b_{j}=0$. $f_{j}$ is then given by the following formula:
\begin{equation}
\label{eq:2}
f_{j}=W^{T}_{j}x = \Vert W_j\Vert \Vert x\Vert cos\theta_{j}
\end{equation}
where $\theta_{j}$ is the angle between $W_{j}$ and $x$. Eq. \ref{eq:2} shows that both the norm and angle of the vector have contribution to the posterior probability.

The constraint without feature normalization is to learns $L2$ norm to minimize the total loss, resulting in relatively weak cosine constraint, which weakens the ability of classification based on cosine similarity. Inspired by cosFace \cite{wang2018cosface}, we normalized the weight vector. Let $\Vert W_{j} \Vert=1$, and fixed $\Vert x \Vert=r$. The representation vector is then mapping on a hypersphere, where $r$ represent the radius and $m$ is the cosine edge at the classification boundary:
\begin{equation}\label{eq:3}
    L_{A} = -\frac{1}{N}\sum^N_{j=1} \log( 
        \frac{
            e^{r_A(\cos\theta_j-m_A)}
        }
        {
            e^{r_A(\cos\theta_j-m_A)}+\sum_{i\not=j}e^{r_A \cos\theta_i}
        }
    )
\end{equation}

Based on Eq. \ref{eq:3}, we introduce domain $D$. In the same domain, features between classes become more widely separated, and the correction loss can be expressed as
\begin{equation}\label{eq:4}
    L_{D} = -\frac{1}{N_{D_i}}\sum^{N_{D_i}}_{j=1,j\in D_i} \log( 
        \frac{
            e^{r_D(\cos\theta_j-m_D)}
        }
        {
            e^{r_D(\cos\theta_j-m_D)}+\sum_{i\not=j}e^{r_D\cos\theta_i}
        }
    )
\end{equation}
$r_A$ and $r_D$ can control the distribution of feature vectors in the hypersphere with different radius, which loosens the feasible space of different domains. $m_D$ is used to control the interval between classes in the domain and adjust the contribution of $L_A$ and $L_D$.

So the total loss $L$ can be formulate as:
\begin{equation}\label{eq:5}
    L = \lambda L_{A} + (1-\lambda) L_D
\end{equation}

\subsection{Intra-/Inter- class separation strategy}

The wide separation of intra-domain and inter-class representation means that more room can be left for the new potential classes. Moreover, learning with more diversity feature representation can help the network learning to be more diverse and transferable. Therefore, we generate a large number of pseudo class data to separate the represented latent space.

We propose a simple and effective method to construct pseudo-class and pseudo-domain. For images, we use the image mixing methods such as Mixup\cite{zhang_mixup_2022}, Cutout\cite{devries2017improved} and CutMix\cite{yun2019cutmix} to randomly sample and fuse in the base session data to generate new and different pseudo-class samples. For a classification task with $C$ class and $D$ domain, we fuse $N_C$ new classes and $N_D$ new domains by random combination. $N_C$ and $N_D$ can be calculated by:
\begin{equation}\label{eq:nc}
    N_C=C\times(C-1)/2
\end{equation}
\begin{equation}\label{eq:nd}
    N_D=D\times(D-1)/2
\end{equation}
The generated pseudo-class and pseudo-domain will be trained together with the original data. In particular, when two data from the same domain are merged, the new data generated still belongs to the source domain. Therefore, the intra-domain category diversity of the source domain will be enhanced.

\begin{table}[t]
    \centering
    \caption{Details of session settings. Two kinds of settings are considered: 1) 1-way 1-shot with 15 incremental sessions, each session learns one category incrementally; 2) Single-domain 1-shot with 3 incremental sessions,  each session categories belong to the same domain.}\label{tab:session_setting}
    \setlength{\tabcolsep}{6mm}{
    \begin{tabular}{cc|cc}
        \hline
        \multicolumn{2}{c|}{1-Way 1-Shot}                                                                          & \multicolumn{2}{c}{Single-domain 1-shot}            \\
        \hline
        \multicolumn{1}{c|}{Session}    & \begin{tabular}[c]{@{}c@{}}Datasets\\ (\# Classes)\end{tabular}          & \multicolumn{1}{c|}{Session} & \begin{tabular}[c]{@{}c@{}}Datasets\\ (\# Classes)\end{tabular} \\
        \hline
        \multicolumn{1}{c|}{1$\sim$5}   & RetinaMNIST(5)                                                          & \multicolumn{1}{c|}{1}       & RetinaMNIST(5)        \\
        \hline
        \multicolumn{1}{c|}{6$\sim$10}  & \begin{tabular}[c]{@{}c@{}}BreastMNIST(2) \\ BloodMNIST(3)\end{tabular} & \multicolumn{1}{c|}{2}       & BreastMNIST(2)        \\
        \hline
        \multicolumn{1}{c|}{11$\sim$15} & BloodMNIST(5)                                                            & \multicolumn{1}{c|}{3}       & BloodMNIST(8)         \\ \hline
    \end{tabular}}
\end{table}
\begin{table}[h]
    \centering
    \caption{Details of subsets.}\label{tab:data_info}
    \setlength{\tabcolsep}{1.8mm}{
    \begin{tabular}{cccccc}
        \hline
        Subset      & Data Modality         & \# Sample & \# Classes & Session\\
        \hline
        PathMNIST\cite{kather2019predicting}   & Colon Pathology       & 107,180   & 9          & Base\\
        DermaMNIST\cite{tschandl2018ham10000}  & Dermatoscope          & 10,015    & 7          & Base\\
        OrganAMNIST\cite{bilic2023liver} & Abdominal CT          & 58,850    & 11         & Base\\
        RetinaMNIST\cite{dataset20202nd}& Fundus Camera         & 1,600     & 5          & Incremental\\
        BreastMNIST\cite{al2020dataset} & Breast Ultrasound     & 780       & 2          & Incremental\\
        BloodMNIST\cite{acevedo2020dataset}  & Blood Cell Microscope & 17,092    & 8          & Incremental\\
        \hline
    \end{tabular}}
\end{table}

\section{Experiment}

\subsection{Dataset of Multiple Diseases}
MedMNIST\cite{yang2021medmnist} is a large standardized biomedical image collection, including 12 number of 2D datasets and 6 number of 3D datasets. All images are pre-propocess to 28 $\times$ 28, with corresponding Two-/multi- class labels.

Our method are experimented on 6 medical disease classification open-source datasets. Following the proposed CDFSCIL, we select PathMNIST\cite{kather2019predicting},DermaMNIST\cite{tschandl2018ham10000}, OrganAMNIST\cite{bilic2023liver} as the base session sets while the others\cite{dataset20202nd,al2020dataset,acevedo2020dataset} are serve as the incremental session. Details are as shown in Table. \ref{tab:session_setting} and Table. \ref{tab:data_info}.

\subsection{Evaluation Criteria}
Following \cite{lifelonger,chaudhry2018efficient}, we report the average accuracy and Performance Dropping rate (PD) for quantitative evaluation. The average accuracy indicates the accuracy of the model after $i$-th session. It can be formulated as $A_t=\frac{1}{t}\sum^t_{i=1} a_{t,i}$, where $a_{t,i}$ is the classification accuracy of model on task $i$ when the training of task $t$ is completed. PD quantifies the decline in the model performance after each task. PD can be calculated as $PD=A_0-A_B$, where $A_0$ indicates accuracy after the base session and $A_B$ indicates the accuracy after the last incremental session. Methods with lower PD suffer less from forgetting.

\subsection{Implementation Details}
For our experiment, we used ResNet18 \cite{he2016deep} as the backbone network. FC uses three layers of fully-connected networks, with 2048 neurons in the first and second layers, and the total number of categories calculated based on Eq. \ref{eq:nc}. The parameters in the loss function are set to 0.8, $r_A=r_D=30$, and $m_A=m_D=0.4$. This method is based on PyTorch library and is optimized with volume driven SGD. The initial learning rate of training was set as 0.01. The Batchsize is 1024. The Limit\cite{limit} and C-FSCIL\cite{hersche2022constrained} configurations are both open source implementations. We performed data enhancement on the sample including random flipping, cropping, and color dithering. In this experiment, mixup was used to generate pseudo-class images.
For each task, we trained the model on an NVIDIA A100 GPU for 100 epochs and chose to stop early when overfitting occurred

Two different experimental configurations were set up to fully evaluate the FSCIL approach, as shown in Table \ref{tab:session_setting}. For 1-way 1-shot, we divide it into 15 sessions, each of which learns one class. Single-domain 1-shot is divided into three sessions, each of which is a separate domain. All Settings are not a training sample class.
\begin{figure}[h]
\includegraphics[width=\textwidth]{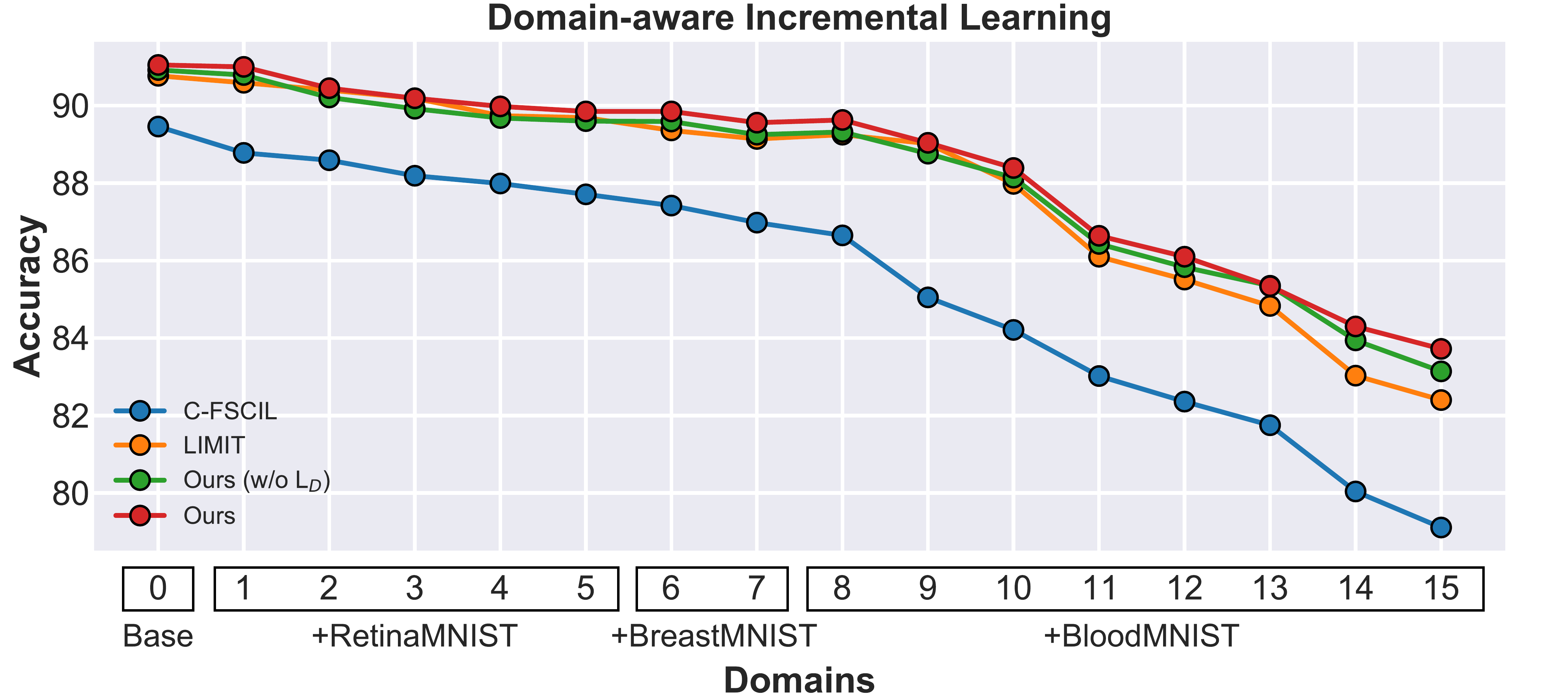}
\caption{Average accuracy under 1-way 1-shot FSCIL setting. Our method is better than other basline methods.} \label{fig:1way}
\end{figure}
\begin{table}
    \centering
    \caption{Average accuracy and PD under single-domain 1-shot.}\label{tab:5way}
    \setlength{\tabcolsep}{3.6mm}{
    \begin{tabular}{cccccc}
        \hline
        Session & 0 & 1 & 2 & 3 & PD $\downarrow$\\
        \hline
        C-FSCIL & 89.46 & 87.76 & 84.00 & 79.21 & 10.25\\
        LIMIT & 90.77 & 89.69 & 87.97 & 82.40 & 8.37\\
        Ours (w/o $L_D$) & 90.92 & 89.60 & 88.14 & 83.14 & 7.79\\
        Ours  & 91.05 & 89.85 & 88.39 & 83.71 & 7.33\\
        \hline
    \end{tabular}}
\end{table}

\subsection{Comparative Results}
As shown in Fig. \ref{fig:1way} and Table. \ref{tab:5way}, our proposed method can maintain the best accuracy and PD under different settings or stages. In Fig. \ref{fig:1way}, As the incremental phase increases, C-FSCIL (blue line) shows a downward trend after $8$-th and $13$-th sessions, respectively. LIMIT (orange line) shows a similar trend to C-FSCIL. Compared to baseline methods, our proposed method shows its superior, especially under large session settings. For the single-domain 1-shot setting, we obtained an average accuracy of 83.71\% in the last session, which is 1.31\% and 4.7\% higher than the LIMIT and C-FSCIL methods, respectively. Particularly, our proposed method is still better in the absence of $L_D$, which means our method could learn various feature representations from pseudo-classes.

\section{Conclusion}
In order to better simulate future clinical scenarios, we first propose a cross-domain few-shot class-incremental learning. In this scenario, in addition to a single medical modality, a complete diagnosis of the patient is required. Our cross-domain incremental learning setting assumes that tasks can come from different medical modalities and datasets. Then, we propose a cross-domain few-shot class-incremental learning method, which improves the feature representation ability of the basic session model by constructing pseudo-sample binding domain enhancement loss, provides a better embedded representation for the incremental categories, and reduces the identification error. Our results show that the performance of cross-domain few-shot class-incremental learning can be improved by properly adjusting the relationships between inter-domain and intra-domain categories in the base session, and our method has substantial improvement over the comparison method.


%
\bibliographystyle{splncs04}
\bibliography{mybibliography}

\end{document}